# The Embodied World Model Based on LLM with Visual Information and Prediction-Oriented Prompts

Wakana Haijima [*,1]  Kou Nakakubo [*,2]
Masahiro Suzuki [3]  Yutaka Matsuo [3]

## Abstract

In recent years, as machine learning, particularly for vision and language understanding, has been improved, research in embedded AI has also evolved. VOYAGER is a well-known LLM-based embodied AI that enables autonomous exploration in the Minecraft world, but it has issues such as underutilization of visual data and insufficient functionality as a world model. In this research, the possibility of utilizing visual data and the function of LLM as a world model were investigated with the aim of improving the performance of embodied AI. The experimental results revealed that LLM can extract necessary information from visual data, and the utilization of the information improves its performance as a world model. It was also suggested that devised prompts could bring out the LLM's function as a world model.

## 1. Introduction

The advancement of machine learning technology has improved the quality of visual recognition and language understanding, leading to increased research activity in embodied AI, where agents with physical presence solve tasks through their actions (Savva, 2019). Among these, the recent development and proliferation of Large Language Models (LLMs) have highlighted the potential of LLM-based embodied AI. A notable example is VOYAGER (Wang, 2023), which operates in the Minecraft environment, continuously exploring the world, acquiring various skills, and making new discoveries without human intervention.

VOYAGER can be regarded as engaging in world model planning; however, LLM-based embodied world models still have issues, including 1) the lack of utilization of visual information and 2) insufficient functionality as world models.

*Equal contribution  [1]University of York, York, United Kingdom  [2] Kyushu Institute of Technology, Fukuoka, Japan  [3] The University of Tokyo, Tokyo, Japan. Correspondence to: Wakana Haijima <wh1229@york.ac.uk>. Copyright 2024 by the author(s).

First, embodied AI is based on the hypothesis that true intelligence emerges from the interaction between an agent and its environment (Smith, 2005). It is known that integrating multiple modalities, such as vision and language, can enhance flexibility and performance (Duan, 2022). The usefulness of image input has also been demonstrated in the planning of embodied world models (Mendonca, 2021). Traditional LLMs like GPT-3 could not utilize visual information as input, but it is considered necessary to incorporate visual information even in LLM-based embodied world models.

In addition, in world models, it is crucial for the agent to simulate various possible events and scenarios in a complex environment and predict future occurrences and responses before selecting actions. Existing world models (Hu, 2023) explicitly program such predictions. On the other hand, it is not clear to what extent LLMs predict future events and responses internally, and this may vary depending on the given prompts. Therefore, to fully leverage LLMs as world models, it is essential to clarify this aspect and enhance their functionality as world models. In this study, to improve the general performance of embodied AI using LLMs, VOYAGER will be extended so that an LLM-based agent could explores Minecraft using multimodal information and prediction-oriented prompts will be applied for enhancing the functionality of LLMs as world models.

## 2. Related Works

### 2.1 Wolrd Model

A world model is a model used by an agent to simulate various possible events and scenarios in the environment and predict future occurrences and responses before choosing actions. In the world model for autonomous vehicle technology (Hu, 2023), the world model virtually simulates the surrounding environment, considering various scenarios the vehicle might encounter and adjusting the vehicle's actions accordingly. Properly modeling the surrounding environment and predicting events is crucial for making flexible and appropriate action choices in complex environments.

Furthermore, an example of an embodied world model utilizing agents with physical presence is LEXA (Mendonca, 2021). LEXA learned a world model from images obtained by exploring the environment and then successfully planned routes to image-based goals, solving tasks in a zero-shot manner. Thus, for an embodied agent to understand its current state and surroundings and function as a world model, the utilization of visual information is essential.

## 2.2 VOYAGER

VOYAGER, using GPT-3.5 and GPT-4, acquired 3.3 times more items, traveled 2.3 times longer distances, and unlocked key tech tree milestones up to 15.3 times faster compared to other LLM-based methods (e.g., ReAct (Yao, 2022), Reflexion (Shinn, 2023) in MineDojo (Fan, 2022)).

One of the main components of VOYAGER, automated curriculum, takes multiple inputs, including the agent's current state (such as the bot's inventory, equipment, health, hunger level, position, nearby blocks and entities, and time), as well as completed and failed tasks. Based on this information, it outputs new task proposals and the reasons for these proposals.

## 3. Proposed Method

### 3.1 Utilizing Visual Information

In VOYAGER, visual information is neither collected nor inputted; instead, it uses Minecraft's cheat information to gather external data for planning. However, to realize an embodied world model, it is necessary to plan based on visual information obtained by the agent, similar to LEXA (Mendonca, 2021), rather than using cheat information. In this study, GPT-4o, which can accept visual information as input, will be utilized to extend VOYAGER, enabling it to use visual information, and the potential of using visual information with LLMs will be evaluated. The following methods for inputting visual data will be compared.

A) Direct use of visual data: A method of directly providing images and other information to GPT-4o.

B) Indirect use of visual data: A method of encoding visual data into text using gpt-4o, then providing the text data to gpt-4o. The following method is used to encode visual data.
  - Free description type: A method for extracting useful information from images to achieve the final goal (*Table 1*)
  - Element extraction type: A method to extract information given by cheats in VOYAGER (Biome, Time, Nearby blocks, Nearby entities) from images.

*Table 1.* Prompt for Converting Visual Information to Text

| Prompt 1. |
|---|
| You are a great assistant to infer information from images. |
| This image is a Minecraft play screen. |
| My ultimate goal is to create a gold pickaxe. |
| Please guess what information you think would be useful to the player and give the information by reading that information from the image. |

### 3.2 Strengthening LLM's Functions as World Model

Prior research indicates that current LLMs have learned spatiotemporal representations of the real world (Gurnee, 2023), and it is possible that VOYAGER also leverages a world model related to Minecraft that the LLM acquired during training. However, VOYAGER currently only instructs the LLM to suggest the next task along with reasons based on the given information, leaving it unclear whether internal future predictions about the environment are being made. To enable LLM-based embodied AI to function as a world model, it is necessary to clarify future predictions utilizing the world model, similar to other existing world models (Hu, 2023).

Therefore, this study proposes using prompt engineering to explicitly instruct the LLM to select the next action based on future predictions. Specifically, the prompts in the automated curriculum will be modified to output both Response1, which suggests the next task under the conventional instructions, and Response2, which explicitly predicts the steps to the goal and changes in the player's state annd the environment after the step before suggesting the next task.

Since only one task can be selected in practice, two patterns will be prepared for comparison: one adopting the task proposed in Response1 (conventional) and the other adopting the task proposed in Response2 (predictive). Performance will be compared between these two patterns.

## 4. Experiments

### 4.1 Experiment Settings

#### 4.1.1 Environmental Settings

The simulation environment used Minecraft, and motor control was handled with Mineflayer. For the LLM, OpenAI's gpt-4-0314 and gpt-3.5-turbo-0301, similar to VOYAGER, were applied and text-embedding-ada-002 was employed for text embedding. However, for experiments with visual information, the multimodal LLM (gpt-4o-2024-05-13) was adopted.

#### 4.1.2 Experimental Design

First, VOYAGER was extended as the description in Section 3.1 and Experiment1 was conducted with the settings of direct image utilization (Experiment 1-1), indirect image utilization in free description format (Experiment 1-2), indirect image utilization in element extraction format (Experiment 1-3), and no image utilization (Experiment 1-4), each performed three times.

Second, VOYAGER was extended as described in Section 3.2 and Experiment2 was conducted with the conventional type (Experiment 2-1) and the virtual type (Experiment 2-2), each performed three times. In contrast to Experiment1, visual information was not utilized here. In Experiment 2-1, cheat information directly obtained from Minecraft was utilized and tasks were proposed by GPT-4, which was same as VOYAGER. Additionally, regardless of whether Response1 or Response2 was adopted, both proposals were outputted in the logs to observe counterfactuals.

All of the above experiments were conducted with the final goal of creating a golden pickaxe, and the maximum number of iterations for task generation was limited to 70.

### 4.1.3 Evaluation Method
When conducting performance comparisons in this experiment, the measurement method employed was the number of iterations for task generation. Additionally, for Experiment 2-1 and Experiment 2-2, the consistency of proposed tasks between Response1 and Response2 was calculated. In this calculation, if tasks of the same type were instructed using items of the same type, it was considered a match. If the types of items or tasks differed, it was considered a mismatch.

## 4.2 Results and Discussion
### 4.2.1 Results for Utilization of Visual Information
The experimental results for Experiments 1-1 to 1-4 regarding the generation process up to the creation of a golden pickaxe are shown in Figure 1. The horizontal axis represents the number of iterations for task generation, while the vertical axis indicates the generation of milestone items. It should be noted that in some cases, the output was "N/A" due to reasons such as unclear images; The image information extraction rate (percentage of outputs other than "N/A") was 97.8% for the free description format (Experiment 1-2) and varied by element for the element extraction format (Experiment 1-3), with 53.4% for Nearby blocks.

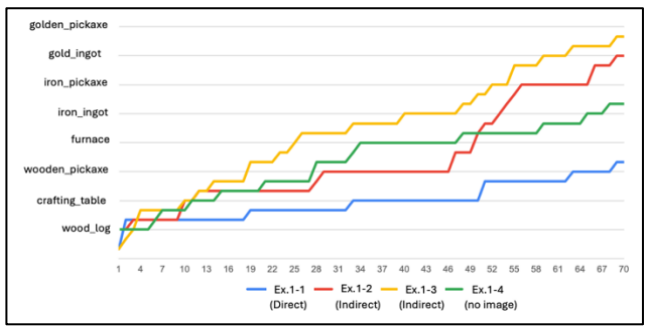

*Figure 1.* Average number of iterations taken to achieve each milestone (Comparison regarding image utilization)

Success in acquiring a golden pickaxe within 70 iterations of task generation was achieved only in the case of indirect utilization of visual data. Specifically, it occurred once out of three trials in the free description format and twice out of three trials in the element extraction format. The average number of task generation iterations until reaching the milestone of acquiring a wooden pickaxe, which is the final goal, was 49.00 iterations for direct image utilization (Experiment 1-1), 22.33 iterations for the free description format (Experiment 1-2), 16.67 iterations for the element extraction format (Experiment 1-3), and 20.67 iterations for no image utilization (Experiment 1-4). It was found that using information extracted from images in the element extraction format allows for faster passage through milestones and achievement of the final goal compared to other methods. This is because information not available from images, such as blocks near the player, can be obtained and utilized. For example, as shown in the examples in *Table 3*, recognition of nearby iron ore from images allowed for planning the acquisition of it as part of the action. Without information from images, it might have been possible to overlook the presence of necessary blocks nearby and continue exploration; By using information extracted from images, more situation-specific action planning could be conducted. From these results, it can be said that effective exploration is possible through the utilization of image information.

*Table 3.* Examples of Planning with Visual Information

| Output 3. Action Ai Message in Experiment 1-3 |
|---|
| Explain: |
| The main issue is that we couldn't find enough iron ore to mine the required amount of raw iron. Since the nearby blocks included iron ore, we should directly mine them instead of continuing to explore. |

On the other hand, when comparing the results of Experiment 1-1 and Experiment 1-4, it was found that direct utilization of images had a rather counterproductive effect on milestone achievement. For example, in the case shown in *Table 4*, despite having an ample amount of wood and a crafting table, thus being in a situation where a wooden pickaxe could be crafted, the suggestion was made to acquire surrounding spruce trees based on the presence of those blocks (wood logs). This can be interpreted as a result of excessive reliance on information from visual sources. However, the observation of such phenomena differs from the results of indirect utilization of images, which also involve recognizing surrounding blocks (Experiments 1-2 and 1-3), thus, it could be said that a multimodal input may influence inference. As indirect utilization of images involves encoding some information from images into text, the difference between indirect and direct utilization encompasses two aspects: the amount of information and the types of modalities. Therefore, with multimodal input, it is conceivable that there may be an increased difficulty in the selection of necessary information for inference due to the abundance of information, or an emphasis on utilizing information from as many modalities as possible when receiving input from multiple modalities.

*Table 4.* Example of Task Proposal with Visual Information

| Output 4. Task Proposal in Experiment 1-1 |
|---|
| Inventory (6/36): {'spruce_planks': 16, 'stick': 8, 'rotten_flesh': 1, 'spruce_log': 5, 'crafting_table': 1, 'spruce_sapling': 1} |
| Reasoning: The main goal is to create a golden pickaxe. One of the essential steps in this process is to have the necessary crafting tools, such as a crafting table and sticks for crafting. Currently, we are in a forested area with spruce trees, which provides an opportunity to gather wood. |
| Task: Obtain a wood log. |

### 4.2.2 Results for Prediction-Orientes Prompts

*Figure 2* shows the generation process up to the golden pickaxe for the results of Experiment 2-1 (conventional type) and Experiment 2-2 (virtual type).

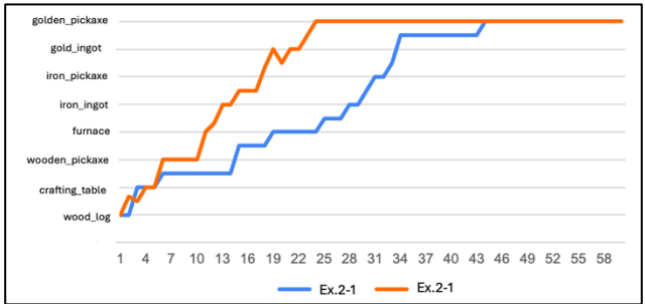

*Figure 2*. Average number of iterations taken to achieve each milestone (Comparison regarding prompts)

From these results, it was found that the prediction-oriented prompt adopted in the virtual type (Experiment 2-2) resulted in fewer task iteration counts for each milestone leading up to the generation of gold ingots. The average number of task iterations required to generate gold ingots was 32.33 in the conventional type and 20.67 in the virtual type, with the virtual type acquiring gold ingots approximately 1.56 times faster. This is believed to be because the virtual type enables more realistic task proposals by clearly predicting the steps to the final goal. As shown in *Table 5*, the conventional type only considered the materials required for acquiring the golden pickaxe and the contents of its own inventory, proposing the generation of gold ingots, which was impossible given the current situation, as the next task, resulting in task failure. In contrast, the virtual type was able to propose the installation of a furnace, necessary for the generation of gold ingots in the pre-step, by considering the steps to acquiring the golden pickaxe, the changes in the situation after each step, and potential risks in executing each step.

From this, it is suggested that by explicitly instructing the roadmap to the final goal and related future predictions through prediction-oriented prompts, the functionality of LLM as a world model can be more effectively leveraged.

*Table 5*. Difference in Task Suggestions

| Output 5. Task Suggestions in Experiment 2-1 |
| --- |
| **Conventional Type** |
| Reasoning: The player has mined gold ore but has not smelted it yet. Since the player's ultimate goal is to create a golden pickaxe, they need to smelt the gold ore to obtain gold ingots. The player has a furnace in their inventory, so they can use it to smelt the gold ore. |
| Task: Smelt 3 raw gold. |
| **Virtual Type** |
| Reasoning: The player has raw gold and sticks in the inventory, and a furnace. The player can smelt the raw gold into gold ingots using the furnace, and then use the gold ingots and sticks to craft a golden pickaxe. |
| Task: Place the furnace. |

Furthermore, the match rate between the proposed tasks in Response1 and Response2 was 22.81% for the conventional type and 35.71% for the virtual type; It was found that by explicitly instructing the steps to the final goal and the subsequent future predictions for each step, more than two-thirds of the proposed tasks would change. This suggests that in the framework of VOYAGER without instructions for future predictions, there may not necessarily be internal future predictions for goal attainment within the LLM.

## 5. Conclusion

In this study, the possibility of utilizing visual information in embodied AI with LLM was verified, and it was attempted to enhance the world model function using prediction-oriented prompts.

The results of the experiments revealed (1) the possibility of effective task proposals by appropriately utilizing encoded visual information, (2) the potential usefulness of prediction-oriented prompting for eliciting the functionality of the world model learned by LLM and improving performance, and (3) the indication that without explicitly instructing future predictions, LLM might not necessarily consider the pathway to the goal internally.

Limitations of this study include being confined to experiments using certain LLMs from the GPT series and conducting experiments focused on a relatively certain task of generating a golden pickaxe as the final goal. As for future work, it would be beneficial to investigate whether similar trends are observed when using other types of LLMs or setting goals with higher uncertainty, such as defeating enemy characters. Additionally, conducting additional controlled experiments, as mentioned in Section 4.2, to examine the impact of multimodal input on inference and to verify the effectiveness of combining the utilization of visual information with prediction-oriented prompting would be valuable.